\documentclass{article}
\usepackage{ijcai17}

\usepackage{times}
\usepackage{algorithm}
\usepackage{algorithmic}
\usepackage{enumerate}
\usepackage{subfigure}
\usepackage{multirow}
\usepackage{epsfig}
\usepackage{graphicx}
\usepackage{amsmath}
\usepackage{amssymb}

\title{Learning with Privileged Information for Multi-Label Classification}
\author{Shiyu Chen, Shangfei Wang\thanks{Shangfei Wang is the corresponding author.}, Tanfang Chen, Xiaoxiao Shi\\
School of Computer Science and Technology, \\University of Science and Technology of China, Hefei, Anhui\\
sy1001@mail.ustc.edu.cn, sfwang@ustc.edu.cn}

\begin{document}

\maketitle

\begin{abstract}
  In this paper, we propose a novel approach for learning multi-label classifiers with the help of privileged information. Specifically, we use similarity constraints to capture the relationship between available information and privileged information, and use ranking constraints to capture the dependencies among multiple labels. By integrating similarity constraints and ranking constraints into the learning process of classifiers, the privileged information and the dependencies among multiple labels are exploited to construct better classifiers during training. A maximum margin classifier is adopted, and an efficient learning algorithm of the proposed method is also developed. We evaluate the proposed method on two applications: multiple object recognition from images with the help of implicit information about object importance conveyed by the list of manually annotated image tags; and multiple facial action unit detection from low-resolution images augmented by high-resolution images. Experimental results demonstrate that the proposed method can effectively take full advantage of privileged information and dependencies among multiple labels for better object recognition and better facial action unit detection.
\end{abstract}
\vspace{-0.4cm}
\section{Introduction}
\begin{figure}[htb]
\centering
\includegraphics[width=0.37\textwidth]{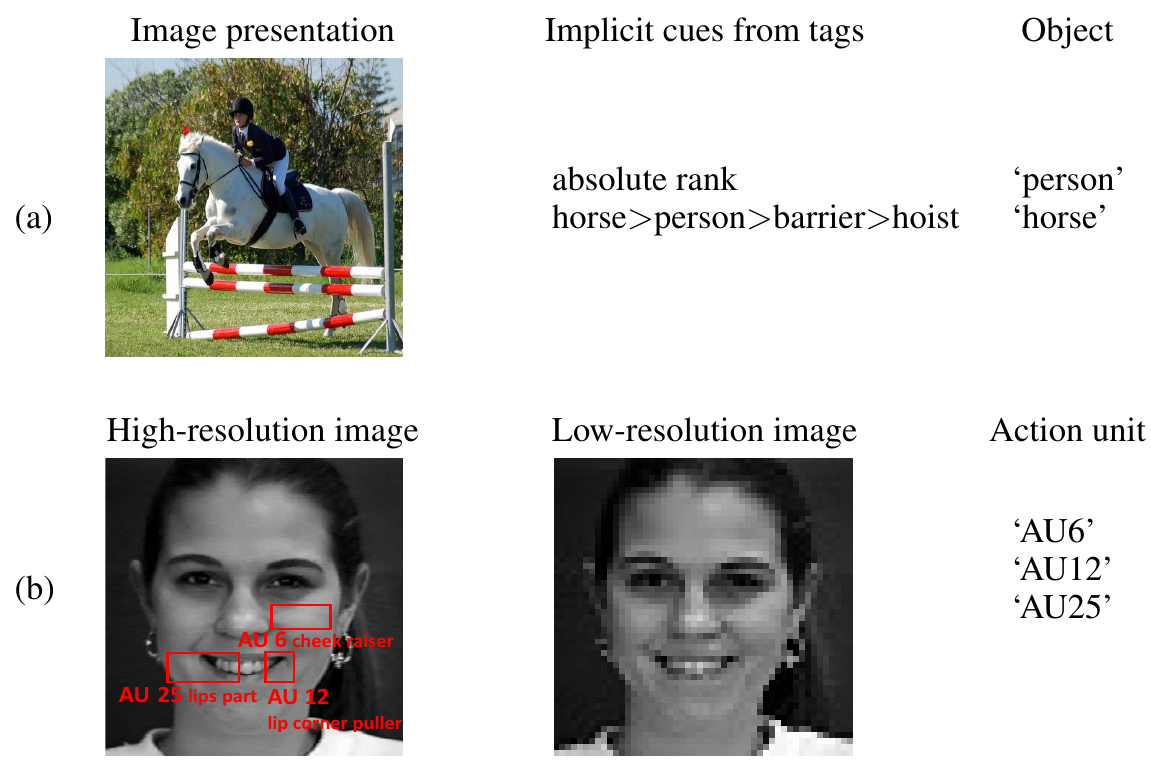}
\caption{Examples of privileged information and multiple labels in different applications.}
\label{fig1} 
\vspace{-0.3cm}
\end{figure}

Traditional supervised learning utilizes the same information during both training and testing. However, in many learning applications, training data have additional information that may be difficult to collect during testing. For example, the order of image tags conveys useful cues about what is most notable to the human viewer. The image shown in Figure~\ref{fig1}, has the following tag list: \{horse, person, barrier, hoist\}. Here, `horse' and `person' occupy the first and second place in the list, and thus have higher absolute rank than `barrier' and `hoist', which appear third and fourth respectively. The implied importance of cues conveyed by the order of manually annotated image tags can be leveraged for object recognition from images during training, but is not available during testing~\cite{Hwang2012Learning}.
As another example, most works use high-resolution facial images for facial action unit detection. However, images which are captured with low-resolution surveillance cameras may be used for action unit detection. The high-resolution images are only available during training.

To address this, Vapnik and Vashist~\cite{Vapnik2009A} proposed a new learning paradigm, i.e., learning using  privileged information~(LUPI), where in addition to available information~(training examples), some privileged information was available during training but not testing. When encoded into the structure or parameters of a classifier, the privileged information can be exploited to construct better classifiers during training. To implement the LUPI paradigm, Vapnik and Vashist~\cite{Vapnik2009A} introduced a discriminative model called Support Vector Machine~(SVM)+, which uses privileged information to predict the slack variables in the SVM.

Since the training of SVM+ is computationally expensive, several  follow-up studies have focused on fast optimization algorithms. For example, instead of using L2-norm as in SVM+, Niu and Wu~\cite{09_niu2012nonlinear} proposed the extended L1-norm SVM for LUPI with nonlinear feature mapping supported by kernel tricks, and  formulated the optimization problem  as linear programming  with less computational cost than the same scale of L2-norm.
In addition to using the privileged data to identify easy- and hard-to-classify examples through slacks, Sharmanska et al.~\cite{sharmanska2013learning} adopted a ranking SVM to identify easy- and hard-to-separate example pairs from privileged information, then transferred the learned ranking prior to the classifier from available information using another ranking SVM. Both ranking SVMs from privileged information and available information are convex optimization problems, and can be solved by standard SVM packages with less computational cost.
Li et al.~\cite{li2016fast} proposed two efficient algorithms for solving the linear and kernel SVM+s. For linear SVM+, they proposed to absorb the bias term into the weight vector, and formulated a new optimization problem with simpler constraints in the dual form.  For kernel SVM+, they applied the L2-loss, and thus obtained a simpler optimization problem in the dual form, with only half of the dual variables of the original SVM+ method.

SVM+ is formulated for binary classification; therefore, several studies extend SVM+ to the multi-class problem. For example,  Liu et al.~\cite{16_liua2013new} formulated the v-K-SVCR method~\cite{17_zhong2006new}, which uses the one-against-rest structure during decomposition, with privileged information to v-K-SCVR+ for multi-class recognition problems. Ji et al.~\cite{19_ji2012multitask} extended the multi-task multi-class support vector machines~($M^2$SVMs) to multi-task multi-class privileged information SVMs~($M^2$PiSVMs) to incorporate the advantages of multi-task learning as well as privileged information. Cai~\cite{21_cai2011advanced} introduced the SVM+-based multi-task learning methods~(SVM+MTL) for classification and regression. Cai~\cite{21_cai2011advanced} also extended generalized Sequential Minimal Optimization~(gSMO) for SVM+MTL.

There are several other approaches in addition to SVM+ and its variants. For example, Chen et al.~\cite{10_chenboosting} developed the Adaboost+ algorithm, which constructs extra weak classifiers from the available information to simulate the performance of better weak classifiers obtained from privileged information.  Yang and Patras~\cite{22_yangprivileged} proposed the privileged information-based conditional regression forest (PI-CRF), which selects the split function based on the information gained from privileged information at some random internal nodes. Wang et al.~\cite{wang2015learning} proposed three-node Bayesian Networks~(BNs) to incorporate privileged information. Wang and Ji~\cite{wang2015classifier} proposed two general approaches to exploit privileged information as additional features and extra labels  to improve different classifiers.
Motiian  et al.~\cite{motiian2016information} extended the information bottleneck principle to LUPI, and expanded it further for learning any type of classifier based on risk minimization.  Pasunuri et al.~\cite{pasunurilearning}  utilized privileged features to create additional labels for training data, and  extended decision trees and boosting for learning with privileged information.
Sharmanska et al.~\cite{sharmanska2014learning} employed privileged information to distinguish between easy and hard examples in the training set, and transferred the information embedded in the privileged information to the available information.
Yan et al.~\cite{yan2016image} proposed an active learning method that uses the learned slack function from privileged information as one of the measurements of uncertainty.
Vrigkas et al.~\cite{vrigkas2016exploiting} introduced a probabilistic classification approach, t-CRF+, to indirectly propagate knowledge from privileged to available feature space through a penalty term , which encourages the model to assign larger weights to samples that have good evidence to distinguish between classes both in privileged and available feature spaces.
Zhang et al.~\cite{zhang2015novel} extended extreme learning machine~(ELM) to ELM+ through relating privileged information to the slack variables.
Yang~\cite{Yang2016Empirical} proposed a metric learning method that jointly learns two distance metrics by minimizing the empirical loss, penalizing the difference between the distance in the original space and that in the privileged space.
Niu et al.~\cite{niu2016exploiting} proposed a new framework called multi-instance learning with privileged information~(MIL-PI) to simultaneously take advantage of privileged information and cope with noise in the loose labels.

All of the above works demonstrate that privileged information can be successfully exploited during training to construct a better binary or multi-class classifier.  However, to the best of our knowledge, little work examines the use of privileged information for multi-label classification.



Multi-label classification has attracted increasing attention in recent years due to its potentially wide applications. For example, as shown in Figure~\ref{fig1}, an image may includes  multiple objects; multiple facial action units may appear on a face.


Due to the large number of possible label sets, multi-label classification is rather challenging. The idea of exploiting the dependencies inherent in multiple labels has been widely employed in the work of multi-label classification. Considering dependencies among labels, current multi-label learning strategies can be categorized into three groups~\cite{zhang2014review}. The first group decomposes the multi-label learning problem into a number of independent binary classification problems, ignoring the dependencies among multiple labels. This method is simple and effective, but may not generalize well due to the ignorance of label correlations. This is referred to as the first order strategy.
The second group considers pairwise relations between labels, such as the ranking between a relevant and irrelevant label, or interaction between any pair of labels. We refer to this method as using second-order strategy.  Since the second group exploits  label correlations to some extent by second-order strategy, it is somewhat generalizable. However, there are certain real-world applications where label correlations go beyond the second-order assumption. The third group considers high-order relations among labels, such as random subsets of the combinations.  A comprehensive survey of multi-label classification types can be found in~\cite{zhang2014review}.

Although current works on multi-label classification successfully leverage label dependencies to recognize multiple labels from an object, almost all multi-label work assumes the same information during both training and testing. There are no existing methods for multi-label classification with the help of privileged information. We incorporate the privileged information and the dependencies among the multiple labels to improve the performance of the multi-label classification.
Specifically, we leverage the extra information-the relationship between available information and privileged information captured by similarity constraints, 
and dependencies among multiple labels captured by ranking constraints. 
During testing, only available information is used. The proposed
method is evaluated on two applications: multiple object recognition from images with the help of implied importance cues embedded in the list of manually annotated image tags; and multiple facial action unit detection from low-resolution images, augmented by high-resolution images. For multiple object recognition, we conduct experiments on the Pascal VOC 2007 database and the LabelMe database. For multiple facial action unit detection, we conduct experiments on the Extended Cohn-Kanade~(CK+) database. {The experimental results demonstrate the relationship between available information and privileged information, and imply that dependencies among multiple labels are beneficial to construct the multi-label classifier. This further indicates the effectiveness of the proposed method.} 



\vspace{-0.1cm}
\section{Problem Statement}
Our goal is to develop a method for multi-label classification with the help of privileged information. During training, both available information and privileged information are utilized to construct a multi-label classifier. During testing, only available information is used.


Given the training data $T = \{(x_i, x^{\star}_i, Y_i )| i=1,...,n\}$, where $x$ represents the available information,  $x^{\star}$ represents the privileged information, $Y$ represents the target labels, and $n$ represents the number of training instances. $Y=\{y_k \in \{-1,1\}|k=1,...,q\}$ indicates the multiple labels, where $q$ represents the number of labels.
The objective of LUPI for multi-label classification is to map the available information of an instance to its multiple labels with the help of privileged information and the label dependencies imbedded in $Y$.  Therefore, the objective function of LUPI for multi-label classification is defined as Eq.~(\ref{solu}).
\vspace{-0.1cm}
\begin{equation}
\begin{aligned}\label{solu}
\min ~L=&\sum_{i=1}^n(\ell(x_i, Y_i)+\ell^{\star}(x_i^{\star},Y_i))+C\sum_{i=1}^nt(x_i,  Y_i)\\
&+C^{\star}\sum_{i=1}^nt^{\star}(x_i^{\star},  Y_i)+D\sum_{i=1}^np(x_i,x_i^{\star},Y_i)
\end{aligned}
\end{equation}
where $\ell(x_i, Y_i)$ and $\ell^{\star}(x_i^{\star},Y_i)$ are the loss functions of the available information classifier and privileged information classifier,  $t(x_i,  Y_i)$ and  $t^{\star}(x_i^{\star},  Y_i)$ reflect the dependencies among multiple labels, and $p(x_i, x_i^{\star}, Y_i)$  reflects the constraints from privileged information.  $C$, $C^{\star}$, and  $D$ are the weighted parameters. Eq.~(\ref{solu}) is a general framework of LUPI for multi-label classification. Therefore, any loss function, constraints from privileged information, and constraints from multi-label dependencies can be used in Eq.~(\ref{solu}).

In this paper, we adopt the maximum margin classifier as the loss function, the similarity between the classifier from available information and that from privileged information as the constraints from privileged information, and the ranking order of the predicted labels as constraints for reflecting multi-label dependencies. 
\vspace{-0.1cm}
\section{Proposed LUPI for Multi-Label Classification}
The privileged information and dependencies among multiple labels are encoded to refine model estimation during training.

Since we use similarity constraints between the classifier from available information and that from privileged information as the constraints from privileged information, we train two classifiers for available information and privileged information simultaneously during training. Thus, we split the training data $T$ into two parts:
\begin{equation}
\begin{aligned}
    TS = \{(x_i, Y_i), (x_i^{\star}, Y_i)|i=1,...,n\}
\end{aligned}
\end{equation}
which can be used to construct classifiers from available information and privileged information, respectively.

For a certain label, two mappings from available information and privileged information can be represented as follows:
\begin{equation}
\begin{aligned}\label{fx}
&f_k({x_i})=<w_k,\phi(x_i)>+b_k,\\
&f^{\star}_k({x^{\star}_i})=<w_k^{\star},\phi^{\star}(x^\star_i)>+b_k^{\star}
\end{aligned}
\end{equation}
where $\phi(x_i)$ and $\phi^{\star}(x^\star_i)$ project $x_i$ and $x^{\star}_i$ into the kernel space.
Then, the  similarity constraint of $\{f_k,f^{\star}_k\}$ is shown as Eq.~(\ref{constraints}):
\begin{equation}
\begin{aligned}\label{constraints}
    |f_k({x_i})-f^{\star}_k({x^{\star}_i})|\leq\eta_{ik} + \epsilon
\end{aligned}
\end{equation}
where $\eta_{ik}$ is the slack variable that measures the failure to meet $\epsilon$ similarity from two classifiers from available information and privileged information.

To exploit the dependencies among multiple labels, we consider the ranking order between present labels and absent labels according to Eq.~(\ref{rank}).
\begin{equation}
\begin{aligned}\label{rank}
    &f_j({x_i})-f_l({x_i})\geq 1-\xi,\\
    &f^{\star}_j({x^{\star}_i})-f^{\star}_l({x^{\star}_i})\geq 1-\xi^{\star}
\end{aligned}
\end{equation}
where $\xi$ and $\xi^{\star}$ are the slack variables to allow some number of disorders that the present labels are ranked below the absent labels. $j$ and $l$ represent the indexes of the present label and absent label respectively.



The similarity constraints from privileged information and the ranking constraints from multiple labels are integrated into the learning process of classifiers from available information. Thus, the objective function of the proposed method is as follows:
\begin{equation}
\begin{split}\label{eq6}
&\mathop{Min}_{\Theta}\sum_{k=1}^q\frac{1}{2}(\Vert{w_k}\Vert^{2}+\Vert{w_k^{\star}}\Vert^{2})+C\sum_{i=1}^n\frac{1}{|Y_i||\overline{Y_i}|}\sum_{(j,l)\in Y_i*\overline{Y_i}}\xi_{ijl}\\
&~~~~~~~~+C^{\star}\sum_{i=1}^n\frac{1}{|Y_i||\overline{Y_i}|}\sum_{(j,l)\in Y_i*\overline{Y_i}}\xi_{ijl}^{\star}+D\sum_{i=1}^n\sum_{k=1}^q\eta_{ik}\\
&\mathbf{s.t.}~~
f_j({x_i})-f_l({x_i})\geq 1-\xi_{ijl},\\
&~~~~~~~~f^{\star}_j({x^{\star}_i})-f^{\star}_l({x^{\star}_i})\geq 1-\xi_{ijl}^{\star},\\
&~~~~~~~~|f_k({x_i})-f^{\star}_k({x^{\star}_i})|\leq\eta_{ik}+\epsilon,\\
&~~~~~~~~\xi_{ik}\geq0,~\xi_{ik}^{\star}\geq0,~\eta_{ik}\geq0\\
&~~~~~~~~all~ for ~(j,l)\in Y_i*\overline{Y_i},~1\leq{i}\leq{n},1\leq{k}\leq{q}\\
\end{split}
\end{equation}
where $\Theta = \{w_k,w_k^{\star},b_k,b_k^{\star}\}$ are the parameters to be optimized, $\{C,C^{\star},D\}$ are the weighted coefficients, and $|Y_i|$ and $|\overline{Y_i}|$ are the numbers of present labels and absent labels for the $i^{th}$ instance. The mappings $f_k(x_i)$ and $f^{\star}_k({x^{\star}_i})$ are represented as Eq.~(\ref{fx}). The first two terms of Eq.~(\ref{eq6}) are the loss functions of the available information classifier and privileged information classifier, respectively. The next two terms are ranking constraints existing in multi-labels. The last term is a slack variable that measures the failure to meet $\epsilon$ similarity from two classifiers from available information and privileged information.

This optimization problem can be translated to its dual problem by applying the Lagrangian multiplier method. The variable $e_{ijl}^k$ represents whether or not the $k^{th}$ label belongs to the label set for $i^{th}$ instance.
\begin{equation}\label{eq:cijlk}
e_{ijl}^k=
\begin{cases}
    +1    &j=k\\
    -1    &l=k
\end{cases}
\end{equation}
where $j$ and $l$ respectively represent the indexes of the present label and absent label for the $i^{th}$ instance.

Thus, the  dual problem is shown as follows:
\begin{equation}
\begin{split}\label{dual}
&\mathop{Max}_{\alpha}-\frac{1}{2}\sum_{k=1}^q\sum_{i,h=1}^n(g_{ik}g_{hk}<x_i,x_h>+g_{ik}^{\star}g_{hk}^{\star}<x^{\star}_i,x^{\star}_h>)\\
&~~~~~~~~~~+\sum_{k=1}^q\sum_{i=1}^n M(\alpha_{ik}+\alpha_{ik}^{\star})\\
&\mathbf{s.t.}~~~~
g_{ik}=\alpha_{ik}\sum_{(j,l)\in{Y_i}*\overline{Y_i}}e_{ijl}^k-(\beta_{ik}^+-\beta_{ik}^-),\\
&~~~~~~~~~~g_{ik}^{\star}=\alpha_{ik}^{\star}\sum_{(j,l)\in{Y_i}*\overline{Y_i}}e_{ijl}^k+(\beta_{ik}^+-\beta_{ik}^-),\\
&~~~~~~~~~\sum_{i=1}^n g_{ik}=\sum_{i=1}^n g_{ik}^{\star}=0,\\
&~~~~~~~~~\alpha_{ik}\in[0,\frac{C}{|Y_i||\overline{Y_i}|}],~\alpha_{ik}^{\star}\in[0,\frac{C^{\star}}{|Y_i||\overline{Y_i}|}],\\
&~~~~~~~~~\beta_{ik}^++\beta_{ik}^-\leq{D},~0\leq{\beta_{ik}^+/\beta_{ik}^-}\\
&~~~~~~~~~all~ for~ (j,l)\in Y_i*\overline{Y_i},~1\leq{i}\leq{n},1\leq{k}\leq{q}\\
&~~~~~~~~~k\in Y_i,M=|Y_i|;~k\in \overline{Y_i},M=|\overline{Y_i}|\\
\end{split}
\end{equation}
where $\alpha=\{\alpha_{ik},~\alpha_{ik}^{\star},~\beta_{ik}^+-\beta_{ik}^-\}$ are the Lagrangian multipliers and $\beta_{ik}^+-\beta_{ik}^-$ serves as a bridge to available information and privileged information. Since Eq.~(\ref{dual}) is a quadratic programming, we adopt the conditional gradient method to solve it.

In the training phase, the similarity constraints from the privileged information and the ranking constraints from the multiple labels are used to construct a better classifier. Since we adopt ranking constraints, the proposed method is a kind of ranking SVM. Therefore, we should train a model to predict the present label set size from training samples, and then use the trained label size predictor to determine the number of present labels of testing samples.

In the testing phase, only available information is used. We use the function $f_k(x_i)$ which is represented in Eq.~(\ref{fx}) to map the features of available information to a real value. Then we assign labels according to the order of mapping values and the number of present labels. Although there is only one feature space during testing, the information from privileged information and multiple labels has been translated and encoded into the classifier parameters.


The detailed algorithm of the proposed method is shown  in Algorithm~\ref{algo1}.
\vspace{-0.2cm}
\begin{algorithm}[htb]
\caption{Algorithm of proposed model}
\label{algo1}
\begin{algorithmic}
\STATE \textbf{Input}:\\
Available information $x$, Privileged information $x^{\star}$,\\
Labels $Y$, Weighted coefficients $\{C, C^{\star}, D\}$\\
\STATE \textbf{Output}: \\
Predicted labels $Z$\\
 \textbf{Stage 1: Predict the mapping values}\\
1)~Choose the kernel function and project $x$ and $x^{\star}$ into the kernel space;\\
2)~Solve the dual problem in Eq.~(\ref{dual}) using the conditional gradient method;\\
3)~Obtain the Lagrangian multiplier $\alpha$, $\alpha^{\star}$, $\beta^+-\beta^-$;\\
4)~Estimate the values of mapping $f$ with the Lagrangian multiplier and projected features from $x$;
\STATE \textbf{Stage 2: Predict label size}\\
5)~Learn a model with the original features and label counts from the training instance;\\
6)~Predict the size of labels of the testing instance, denoted as $\psi$;\\
\STATE \textbf{Stage 3: Obtain the predicted labels}\\
7)~Output the predicted labels $Z$ by considering a label $k$ is the label set of $x$ if the corresponding mapping value is among the top $\psi$ elements.
\end{algorithmic}
\end{algorithm}
\vspace{-0.2cm}
\section{Experiments}
\subsection{Experimental Conditions}
We evaluate the proposed method on two applications: multiple object recognition from images with the help of implicit importance cues from the order of image tags supplied by annotators, and multiple action unit detection from low-resolution facial images enhanced by high-resolution facial images. Two image benchmark databases~(the Pascal VOC 2007 database~\cite{everingham2010pascal} and the LabelMe database~\cite{russell2008labelme}), and one expression database~(the CK+ database~\cite{Lucey2010The}) are adopted.

\begin{table*}
\centering
\caption{Experimental results on the Pascal VOC 2007 database, the LabelMe database, and the CK+ database.} 
\label{table1}
\newcommand{\tabincell}[2]{\begin{tabular}{@{}#1@{}}#2\end{tabular}}
\begin{tabular}{c|cccc}
\hline
Database&SVM&SVM+ML&SVM+PI&Ours\\
\hline
{Pascal VOC 2007}
 &0.252$\backslash$0.287$\backslash$0.155&0.315$\backslash$0.368$\backslash$0.186&0.300$\backslash$0.348$\backslash$0.169&\textbf{0.346}$\backslash$\textbf{0.400}$\backslash$\textbf{0.211}\\
 \hline
 {LabelMe}
 &0.548$\backslash$0.655$\backslash$0.154&0.602$\backslash$0.699$\backslash$0.220&0.561$\backslash$0.670$\backslash$0.158&\textbf{0.606}$\backslash$\textbf{0.704}$\backslash$\textbf{0.223}\\
 \hline
 \multirow{1}{1.3cm}{CK+}
 &0.555$\backslash$0.646$\backslash$0.268&0.609$\backslash$0.694$\backslash$0.353&0.575$\backslash$0.668$\backslash$0.280&\textbf{0.621}$\backslash$\textbf{0.707}$\backslash$\textbf{0.358}\\
 \hline
\end{tabular}

"-$\backslash$-$\backslash$-" refers to example-based accuracy, example-based F-measure, and example-based subset-accuracy respectively.
\vspace{-0.1cm}
\end{table*}
The Pascal VOC 2007 database consists of 9963 samples of images and 20 target labels. For image representation, 200 dimensional bag of visual word features, 512 dimensional gist features~\cite{oliva2006building}, and 64 dimensional color histogram features provided by~\cite{hwang2010accounting} are used. For implicit importance cues from the order of image tags supplied by annotators, 339 dimensional absolute tag rank features provided by~\cite{hwang2010accounting}~\cite{Hwang2012Learning} are used.  The train-test split provided by the database constructor is adopted~(5011 trainning and 4952 testing samples).

The LabelMe database consists of 3852 samples of images and 209 target labels. Like the Pascal VOC 2007 database, for image presentation, 200 dimensional bag of visual word features, 512 dimensional gist features~\cite{oliva2006building}, and 64 dimensional color histogram features provided by~\cite{hwang2010accounting} are employed. For implicit importance cues from the order of image tags supplied by annotators, 209 dimensional absolute tag rank features provided by~\cite{hwang2010accounting}~\cite{Hwang2012Learning} are used.  Most of the target labels are sparse; therefore, labels representing less than 15\% of all samples are discarded.  A random 50-50 split is performed on the database to construct training and testing sets. 1889 instances with 14 labels are obtained for training, and 1878 instances are obtained for testing.

The CK+ database includes 593 facial expression image sequences staring from the neutral frame and ending with the peak frame~\cite{Lucey2010The}.  The original images, with resolutions of  640*490, are viewed as high-resolution images. We generate  low-resolution images by normalizing the face region to a 256*256  image. For high-resolution images, the similarity-normalized shape differences between the neutral frame and exaggerated frame are used as features~\cite{Lucey2010The}. The similarity-normalized shape differences are difficult to obtain for low-resolution images, so local binary pattern features~\cite{Ahonen2006Face} are extracted from the difference images. PCA is used for dimension reduction. 
Facial action units available for more than 15\% of all samples are chosen.  Thus, we obtain  575 samples with 9 labels.  We perform 10-fold subject independent cross validation.

%

For multiple object recognition from images with the help of implicit importance cues from the order of image tags, we use image representation as available information, and implicit importance cues from the order of image tags as privileged information. For multiple action unit detection from low-resolution facial images enhanced by high-resolution facial images, we use low-resolution images as the available information, and high-resolution images as the privileged information.

To evaluate the performance of the proposed method, we conduct four experiments: one using only available information~(SVM), one using available information with the help of the dependencies among multiple labels~(SVM+ML), one using available information with the help of the privileged information~(SVM+PI), and the proposed method~(Ours). Example-based accuracy, example-based F-measure, and example-based subset-accuracy are used as performance metrics~\cite{sorower2010literature}.

\subsection{Experimental Results and Analysis}

Table~\ref{table1} shows the experimental results on the Pascal VOC 2007 database, the LabelMe database, and the CK+ database. From Table~\ref{table1}, we observe the following:

First, SVM+ML achieves higher accuracy, F-measure, and subset-accuracy than SVM on all three databases.
Compared to SVM, SVM+ML takes advantage of the dependencies among multiple labels. The improvement is especially obvious on the LabelMe database and the CK+ database. This may be caused by the more balanced label distribution of these databases, which may make label dependencies easier to capture. Specifically, for the Pascal VOC 2007 database, just 10\% of samples have more than two present labels, while for the LabelMe database and the CK+ database, 70\% and 56\% samples have at least three present labels, respectively. This indicates the importance of label dependencies for multi-label classification.

Second, compared to SVM, SVM+PI achieves better performance in terms of accuracy, F-measure, and subset-accuracy on all three databases. SVM is a traditional supervised learning method which utilizes the same information during both training and testing. SVM+PI captures and exploits the relationships between the available information and privileged information during training. This demonstrates that the privileged information can be exploited during training to build better classifiers.

Third, SVM+ML achieves better performance than SVM+PI for all metrics on all three databases. SVM+ML and SVM+PI use either label dependencies and privileged information, which are both crucial for multiple classification. Unlike the privileged information, which is only available during training, but unavailable during testing, label dependencies not only exist in the ground-truth labels during training, but also inhere in the predicted labels during testing, since multiple objects and multiple AUs coexist on a scene image and a facial image respectively. This indicates that using the dependencies among multiple labels builds a better classifier than one built using privileged information.

Finally, the proposed method performs best on all three databases. The proposed method not only considers the relationship between available information and privileged information, but also considers the dependencies among multiple labels. This demonstrates that the proposed method can successfully captures  the privileged information and label dependencies for multiple classification.

 To further demonstrate the importance of privileged information and label dependencies, we analyze the recognition performance of every  label on the LabelMe database and the CK+ database. Table~\ref{table4} and Table~\ref{table2} show the F-measure for each label on the LabelMe database and the CK+ database, respectively. From Table~\ref{table4} and Table~\ref{table2}, we find that:
\begin{table}
\centering
\caption{F-measure for each object on the LabelMe database.} 
\label{table4}
\newcommand{\tabincell}[2]{\begin{tabular}{@{}#1@{}}#2\end{tabular}}
\begin{tabular}{c|cccc}
\hline
Object&SVM&SVM+ML&SVM+PI&Ours\\
\hline
`unmatched'&0.533&0.585&0.517&\textbf{0.586}\\
`person'&0.655&0.662&0.634&{0.637}\\
`window'&0.492&0.602&0.528&\textbf{0.634}\\
`car side'&0.530&0.574&0.599&\textbf{0.613}\\
`tree'&0.625&0.698&0.641&{0.697}\\
`building'&0.800&0.873&0.800&\textbf{0.875}\\
`sky'&0.725&0.722&0.776&{0.750}\\
`road'&0.753&0.872&0.784&0.862\\
`sidewalk'&0.606&0.770&0.671&0.765\\
`sign'&0.305&0.463&0.551&\textbf{0.551}\\
`table'&0.831&0.809&0.821&{0.798}\\
`screen'&0.901&0.857&0.900&{0.867}\\
`keyboard'&0.721&0.729&0.7291&\textbf{0.777}\\
`car'&0.746&0.787&0.767&0.782\\
 \hline
 Avg.&0.659&0.715&0.694&\textbf{0.728}\\
 \hline
\end{tabular}
\vspace{-0.4cm}
\end{table}
\begin{table}
\centering
\caption{F-measure for each action unit on the CK+ database.} 
\label{table2}
\newcommand{\tabincell}[2]{\begin{tabular}{@{}#1@{}}#2\end{tabular}}
\begin{tabular}{c|cccc}
\hline
Action unit&SVM&SVM+ML&SVM+PI&Ours\\
\hline
AU1&0.603&0.644&0.605&\textbf{0.661}\\
AU2&0.611&0.688&0.661&\textbf{0.711}\\
AU4&0.596&0.624&0.624&\textbf{0.655}\\
AU5&0.604&0.677&0.598&0.667\\
AU6&0.447&0.506&0.487&\textbf{0.508}\\
AU7&0.474&0.508&0.494&\textbf{0.547}\\
AU12&0.603&0.619&0.657&\textbf{0.674}\\
AU17&0.742&0.746&0.754&0.742\\
AU25&0.833&0.825&0.846&0.828\\
 \hline
 Avg.&0.613&0.648&0.636&\textbf{0.666}\\
 \hline
\end{tabular}
\vspace{-0.4cm}
\end{table}

 Compared to SVM, SVM+ML achieves a better average F-measure and significantly improves on several labels, such as `building' and `road' on the LabelMe database and AU1 and AU2 on the CK+ database. The improvements are about 7\% and 12\% for `building' and `road', and 4\% and 7\% for AU1 and AU2 respectively. Through analyzing the ground truth labels, we find that $P('building'=1|'road'=1)=84.34\%,P('road'=1|'building'=1)=80.53\%$ and $P(AU1=1|AU2=1)=100\%,P(AU1=2|AU1=1)=66.86\%$, showing high positive correlations between `building' and `road', and between AU1~(inner brow raiser) and AU2~(outer brow raiser). Such coexistence is successfully captured by the proposed ranking constraints, and results in better performance.

 Compared to SVM, SVM+PI has a better average F-measure on both databases. The most significant improvement is about 24\% for `sign' on the LabelMe database. The image features capture the total scene structure, color, and the appearance of component objects. However, due to the diversity of `sign', it is difficult to recognize from image features. The implicit importance cues from the order of image tags supplied by annotators may capture the importance of `sign', making it easier to recognize. Similarly, the most significant improvement on the CK+ database is about 5\% for AU12. Since AU12 captures the lip corner puller, it is more difficult to detect from low-resolution facial images. High-resolution images are more helpful for AU12 detection. This demonstrates that the proposed method successfully leverages  privileged information to construct better classifiers during training, especially for those objects which are difficult to recognize from available information only.

 The proposed method achieves the best results on the average F-measure on the two databases. This indicates that both privileged information and label dependencies contribute to better multi-label classification.

\vspace{-0.2cm}
\subsection{Comparison to Related Work}
Since current works on LUPI do not consider multi-label classification, a direct comparison cannot be made. We compare our work to a related method, SVM+~\cite{Vapnik2009A}\cite{li2016fast}. 
SVM+ is used to recognize objects from images with the help of implicit importance cues from the order of image tags supplied by annotators on the Pascal VOC 2007 database and the LabelMe database, and to detect action units from low-resolution images enhanced by high-resolution images on the CK+ database.
\begin{table}
\footnotesize
\caption{Results of SVM+ \protect\cite{li2016fast} on the three databases.}
\label{table3}
\newcommand{\tabincell}[2]{\begin{tabular}{@{}#1@{}}#2\end{tabular}}
\begin{tabular}{c|ccc}
\hline
Database&\tabincell{c}{Example-based\\accuracy}&\tabincell{c}{Example-based \\F-measure}&\tabincell{c}{Example-based \\subset-accuracy}\\
\hline
VOC&0.321&0.366&0.197\\
LabelMe&0.555&0.648&0.249\\
CK+&0.566&0.662&0.249\\
\hline
\end{tabular}
\vspace{-0.5cm}
\end{table}

The experimental results of SVM+ on the three databases are illustrated in Table~\ref{table3}. {Table~\ref{table1} and Table~\ref{table3} show that:

First, since SVM+ considers the relationship between available information and privileged information, it has better results than SVM, with higher performance metrics on all three databases. It indicates the effectiveness of privileged information.

Second, SVM+PI and SVM+ have similar results. Specifically, SVM+PI has higher accuracies and F-measures on the three databases, while SVM+ has higher subset-accuracies on two databases. SVM+ assumes that privileged information and available information share the same slacking variable, and SVM+PI adopts similarity constraints on the classifiers from privileged and available information. The two methods have similar abilities to capture privileged information.


Finally, the proposed method outperforms SVM+ in all performance metrics on the three databases. Compared to SVM+, the proposed method not only considers privileged information, but also utilizes the multi-label dependencies to refine the parameters of the classifier. The experimental results demonstrate the usefulness of the dependencies among multiple labels for multi-label classification.

\vspace{-0.2cm}
\section{Conclusions}
In this paper, we propose a new multi-label classification method from available information with the help of privileged information. Specifically, we adopt the similarity constraints between the available information and privileged information to adjust the performance of the classifier. We also utilize the ranking constraints from the dependencies among multiple labels to improve classification performance. The proposed method is evaluated on different applications. Experimental results on three benchmark databases show that both privileged information and dependencies among multiple labels are beneficial for building a better classifier. This further demonstrates the effectiveness and superiority of the proposed method as compared to state-of-the art methods.
{

\bibliographystyle{named}
\small
\bibliography{ijcai17}
}
\end{document}